\documentclass[journal,twoside,web]{ieeecolor}
\usepackage{tmi}
\usepackage{cite} 
\usepackage{amsmath,amssymb,amsfonts}
\usepackage{graphicx}
\usepackage{textcomp}
\usepackage{multirow}
\usepackage{bm}
\usepackage{algorithm}
\usepackage{algpseudocode}
\usepackage{arydshln}
\usepackage{afterpage}
\usepackage{booktabs}

\def\BibTeX{{\rm B\kern-.05em{\sc i\kern-.025em b}\kern-.08em
    T\kern-.1667em\lower.7ex\hbox{E}\kern-.125emX}}
\begin{document}
\title{SAC-MIL:  Spatial-Aware Correlated Multiple Instance Learning for Histopathology Whole Slide Image Classification}
\author{
Yu Bai, 
Zitong Yu, 
Haowen Tian,
Xijing Wang,
Shuo Yan,
Lin Wang,
Honglin Li,
Xitong Ling,
Bo Zhang,
Zheng Zhang,
Wufan Wang,
Hui Gao,
Xiangyang Gong,
Wendong Wang,
\IEEEmembership{Member, IEEE}
\thanks{This paragraph of the first footnote will contain the date on which
you submitted your paper for review. It will also contain support information,
including sponsor and financial support acknowledgment. For example, 
``This work was supported in part by the U.S. Department of Commerce under Grant BS123456.'' }
\thanks{The next few paragraphs should contain the authors' current affiliations,
including current address and e-mail. For example, F. A. Author is with the
National Institute of Standards and Technology, Boulder, CO 80305 USA (e-mail:author@boulder.nist.gov). }
\thanks{S. B. Author, Jr., was with Rice University, Houston, TX 77005 USA.
He is now with the Department of Physics, Colorado State University,
Fort Collins, CO 80523 USA (e-mail: author@lamar.colostate.edu).}
\thanks{T. C. Author is with the Electrical Engineering Department,
University of Colorado, Boulder, CO 80309 USA, on leave from the National
Research Institute for Metals, Tsukuba, Japan (e-mail: author@nrim.go.jp).}
}

\maketitle

\begin{abstract}
We propose Spatial-Aware Correlated Multiple Instance Learning (SAC-MIL) for performing WSI classification.
SAC-MIL consists of a positional encoding module to encode position information and a SAC block to perform full instance correlations.
The positional encoding module utilizes the instance coordinates within the slide to encode the spatial relationships instead of the instance index in the input WSI sequence.
The positional encoding module can also handle the length extrapolation issue where the training and testing sequences have different lengths.
The SAC block is an MLP-based method that performs full instance correlation in linear time complexity with respect to the sequence length.
Due to the simple structure of MLP, it is easy to deploy since it does not require custom CUDA kernels, compared to Transformer-based methods for WSI classification.
SAC-MIL has achieved state-of-the-art performance on the CAMELYON-16, TCGA-LUNG, and TCGA-BRAC datasets.
The code will be released upon acceptance.
\end{abstract}



\begin{figure}[t]
\vspace{-1.0em}
    \centering
    \includegraphics[width=1\linewidth]{"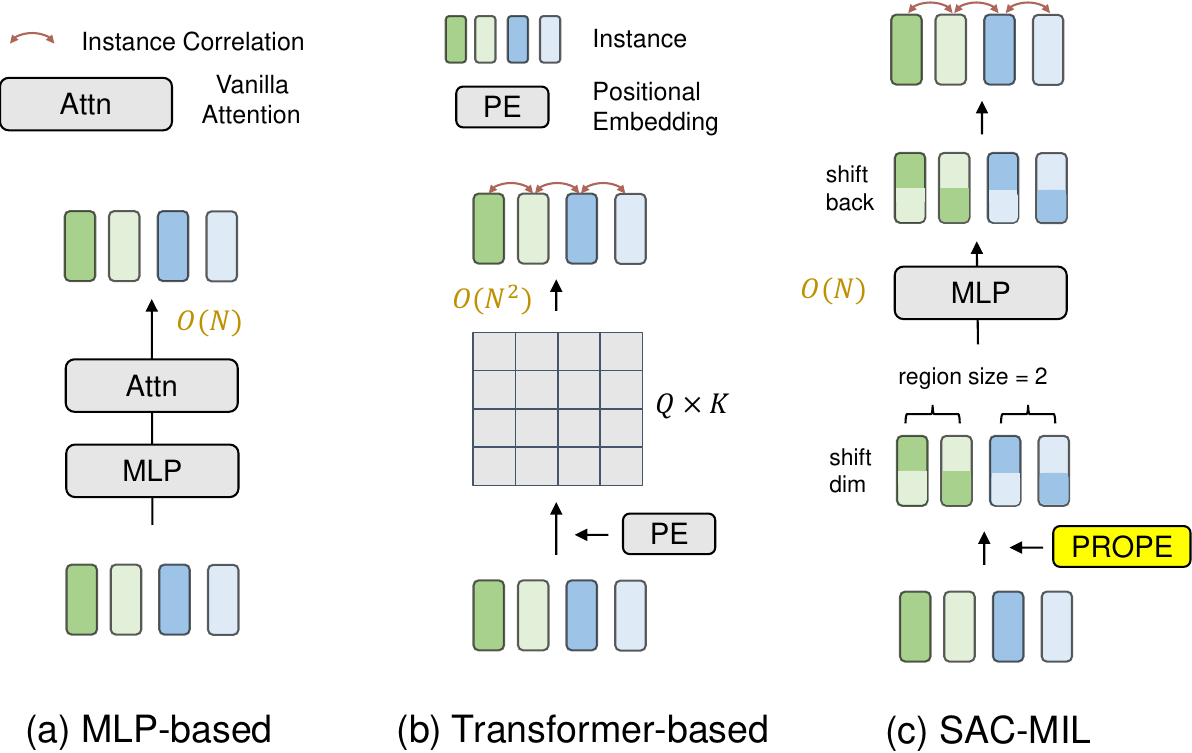"}
    \vspace{-1.3em}
    \caption{
    (a) MLP-based methods employ MLPs with vanilla attention (e.g., Sigmoid function) to perform WSI classification. These methods exhibit linear complexity with respect to the instance number but do not explicitly model correlations between instances.
    (b) Transformer-based methods utilize self-attention mechanisms with quadratic complexity to capture correlations between instances and also encode positional information for each instance.
    (c) SAC-MIL encodes positional information and performs full instance correlations using MLP layers through channel shifting.
    For simplicity, not all instance correlations are illustrated.
        }
    \label{fig:fig1}
    \vspace{-1.5em}
\end{figure}

\vspace{-1.3em}
\section{Introduction}
\vspace{-0.3em}
\label{sec:introduction}
Whole Slide Image (WSI) classification plays a critical role in computer-aided diagnosis (CAD)~\cite{shapely}.
%
WSI classification aims to perform tumor classification of a given slide. 
However, each slide exhibits an extremely high resolution of approximately $100,000 \times 100,000$ pixels, making the annotation of tumor regions a time-consuming process.
To reduce the dependency on detailed tumor region annotation, recent research commonly adopts Multiple Instance Learning (MIL) approach for Whole Slide Image (WSI) classification~\cite{dsmil, shapely, dtfd, transmil, clam, hipt}.
Under the MIL setting, each slide is treated as a "bag," and the individual patches within the slide are considered "instances." 
The objective is to perform bag-level classification based on instance-level inputs while utilizing only the bag-level labels.
Owing to this advantage, MIL-based approaches have become dominant for WSI classification.


MIL-based methods can be broadly categorized into two primary types: \textbf{Multi-Layer Perceptron (MLP)-based} and \textbf{Transformer-based} approaches, as illustrated in Fig.~\ref{fig:fig1}.
\textbf{MLP-based} methods~\cite{dtfd, clam, dsmil} employ MLPs augmented with vanilla attention to process all instances simultaneously, achieving linear complexity.
Thus, MLP-based methods can efficiently process a large number of instances at a low computational cost, making them particularly suitable for WSI classification tasks where the number of instances often exceeds 10,000~\cite{transmil}.
However, MLP-based methods generally lack the ability to model full correlations between instances and often ignore the positional information of each instance.
To alleviate these issue, \textbf{Transformer-based} methods~\cite{transmil, hipt, re_embedding, norma} utilize the Transformer architecture~\cite{transformer, vit} to model full instance correlations and encode positional information through a positional encoding module.

However, Transformer-based methods still face certain shortcomings in WSI classification.
In WSI classification, the input sequence can be long due to the high resolution of WSIs. 
When handling such long sequences, the Transformer architecture suffers from  high computational cost~\cite{transformer, vit} .
To reduce computational cost, recent Transformer-based methods have adopted various efficient self-attention mechanisms~\cite{gigapath, transmil, norma, hipt}.
However, custom CUDA kernels require specific GPU architectures, thus limits the deployment under the GPU shortage in clinical settings~\cite{hajdu2021covid_short, fernandes2024nvidia_short}.
However, these efficient self-attention mechanisms still face two main challenges:
\textbf{(1) Performance Degradation}: Some methods~\cite{transmil, Nystrom, gigapath, longnet, transformerxl} compute approximated self-attention instead of full self-attention~\cite{transformer}, which may lead to performance degradation.
\textbf{(2) Hardware Compatibility}: Other methods~\cite{flash} compute full self-attention using custom CUDA kernels. 
However, custom CUDA kernels require specific GPU architectures, thus limiting the deployment under the GPU shortage in clinical settings~\cite{hajdu2021covid_short, fernandes2024nvidia_short}.

On the contrary, MLP-based methods do not face the challenges present in Transformer-based methods.
Since MLP-based methods exhibit linear computational complexity with respect to the sequence length, MLP layers can be easily adopted to process long sequences without modification. 
And the simple structure of MLP layers also makes them easy to deploy.
However, MLP-based methods lack the ability to encode positional information and perform full instance correlation compared to Transformer-based methods.
Therefore, we pose the following question: \textit{Can we enable MLP-based methods to encode positional information and perform full instance correlations so that they can have the advantages of Transformer-based methods while eliminating their disadvantages?}

\begin{figure*}[t]
\vspace{-1.6em}
    \centering
    \includegraphics[width=0.9\textwidth]{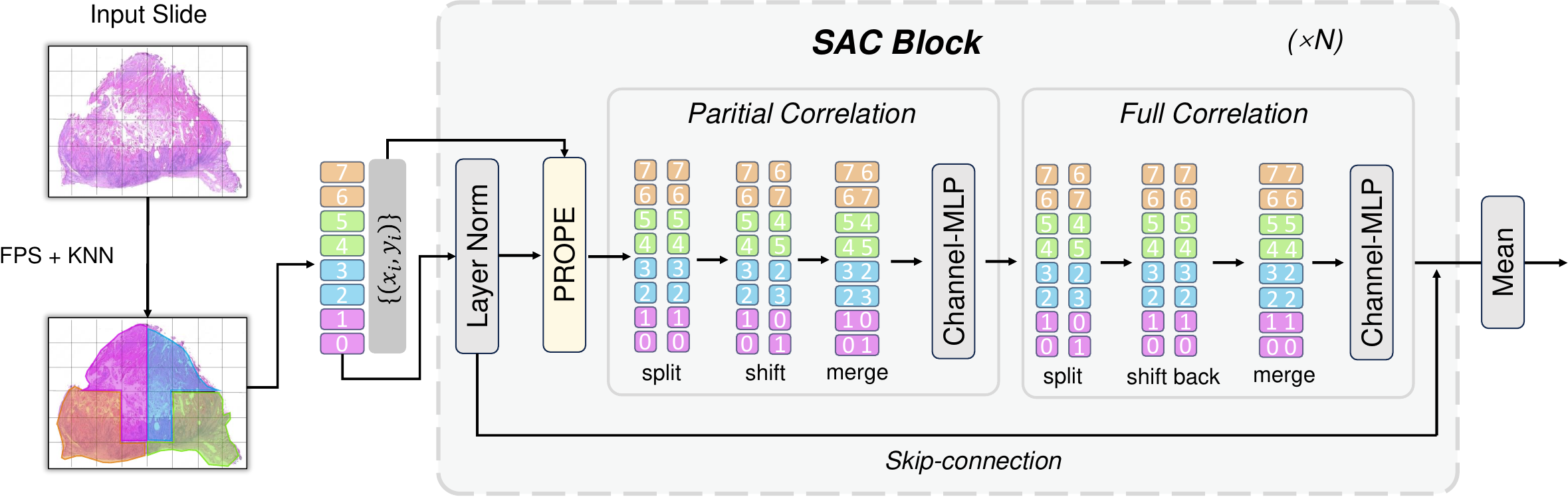}
    \vspace{-0.6em}
    \caption{
        Overall Architecture of SAC-MIL. 
        SAC-MIL uses FPS and KNN to split instances into multiple regions, indicated by different colors.
        PROPE encodes positional information based on spatial coordinates.
        Each SAC block first splits each instance into multiple folds along the channel dimension, shifts each fold with increased steps, and finally applies a channel-wise MLP to perform partial instance correlations.
        It then performs full instance correlation by shifting the folds back to their original positions within the same region.
        After each SAC block, all instances within the same region are fully correlated.
        The region size increases exponentially as the number of SAC blocks increases.
    }
    \label{fig:fig2}
    \vspace{-1.2em}
\end{figure*}

To achieve this goal, we propose Spatial-Aware Correlated Multiple Instance Learning (SAC-MIL) for WSI classification.
The overall architecture of SAC-MIL is illustrated in Fig.~\ref{fig:fig2}.
We first divide all instances into multiple regions using FPS~\cite{FPS} (Farthest Point Sampling) and KNN (K-Nearest Neighbors) to encode the local context.
Then, Polar Rotary Position Embedding (PROPE) is introduced to encode the positional information of each instance.
Finally, a stack of SAC blocks are used to gradually perform instance correlations, inspired by~\cite{chord_mixer, cycle_mlp}.
PROPE adopts spatial coordinates of each instance within the WSI instead of the token index.
The token index cannot fully reflect the spatial information of each instance due to the irregular shapes of tissue areas.
Additionally, PROPE adopts the polar coordinate system instead of the traditional Cartesian coordinate system~\cite{2d_rope_fit, 2d_rope_unified} to encode the positional information of WSIs.
This approach ensures that each dimension can obtain full positional information, rather than partial positional information (single axial) as in 2D ROPE~\cite{2d_rope_fit, 2d_rope_unified, rope_eccv}.
Furthermore, PROPE normalizes the polar coordinates of both training and testing samples to the same range, alleviating the length extrapolation issue~\cite{PI} when both sequences have different lengths.
 SAC blocks adopt a two-step process to perform instance correlations within each region:
(1) Partial correlation: Split all instances into multiple folds, shift each fold with an increased step, and perform fold-level correlations using a channel-wise MLP.
(2) Full correlation: Reverse each fold to its original position and apply another channel-wise MLP to process sequences of various lengths.
The first step performs fold-level correlations so that each fold contains partial features of other instances from the same region.
The second step performs instance-level correlations, as each fold now contains all the features of other folds within the same region.
After these two steps, full instance correlation is performed within each region.
The region size increases exponentially as the number of layers increases.
Therefore, the Effective Context Length (ECL) can quickly reach the maximum length.
SAC-MIL can encode positional information and perform full instance correlation with linear time complexity. 
Its simple MLP-style architecture also facilitates easy deployment.

The contributions of SAC-MIL are as follows:
(1) We propose SAC-MIL to model instance correlations with linear computational complexity and encode positional embeddings. 
(2) We introduce a positional embedding layer named PROPE to encode the positional information of each instance. PROPE uses a polar coordinate system as the token index and normalized patch coordinates to better model the spatial relationships between instances.
(3) We propose a SAC block to model full instance correlations. The SAC-MIL progressively encodes local context as the region size increases and effectively handles input sequences of various lengths.

\section{Related Work}

\noindent \textbf{WSI classification}
There are mainly two categories of WSI classification tasks: MLP-based methods and Transformer-based methods. 
MLP-based methods~\cite{clam, dsmil, dtfd} primarily adopt an MLP layer with vanilla attention to perform WSI classification tasks.
AB-MIL~\cite{abmil} computes vanilla attention for each instance and aggregates all instances to form the bag-level representation, which is then used for prediction.
CLAM~\cite{clam} extends the MIL classification task to a multi-class setting.
DS-MIL~\cite{dsmil} uses an MLP to reduce each instance to a scalar as the instance score and computes the cosine similarities between the highest instance score and all other instances to perform partial interactions among the top-k instances and other instances.
Although MLP-based methods have achieved promising results, they often lack full instance correlations and ignore instance position information.
Transformer-based methods~\cite{norma, transmil, gigapath, hipt} adopt the Transformer architecture~\cite{vit, transformer} to compute instance interactions.
Trans-MIL~\cite{transmil} adopts Nystrom attention~\cite{Nystrom} to perform instance correlations.
Gigapath~\cite{gigapath} adopts LongNet~\cite{longnet} to perform instance interactions, where LongNet uses dilated self-attention to reduce computational cost.
Transformer-based methods often adopt efficient self-attention mechanisms to model full self-attention~\cite{vit, transformer}.
Some methods adopt Mamba~\cite{mambamil, mammil} architecture to reduce the computational cost.
However, this may result in suboptimal performance~\cite{chord_mixer, s4, sparse} or require custom CUDA kernels~\cite{longnet, flash}.

\noindent \textbf{Positional Encoding}
Transformer-based methods have adopted various positional encoding methods~\cite{transmil, re_embedding, gigapath, norma} for WSI classification.
These methods use token indices to encode positional information.
The token index in the WSI sequence does not fully reflect the spatial relationship between instances in the WSI when tokens are flattened into a sequence, especially because the shape of the tissue area is irregular.
However, the irregular shape will cause length extrapolation issues~\cite{alibi, PI} for ROPE-based methods.
PI~\cite{PI} alleviates this issue by normalizing the token index during inference to the same length as the training sequence.
However, this approach can only handle situations where the length of inference sequences is greater than that of the training sequences.
ROPE 2D is often used in vision tasks~\cite{2d_rope_fit, 2d_rope_unified}.
However, it only encodes partial positional information (one axial) for each dimension, which may result in a lack of complete positional information.

\noindent \textbf{MLP Mixer}
MLP-Mixer~\cite{mlpmixer, gmlp, ViP, cycle_mlp, chord_mixer, resmlp, Wave-MLP, GFNet} is an MLP-like architecture that uses only MLP layers to perform full instance interactions instead of self-attention.
However, these methods cannot handle input sequences of various lengths.
Chord-Mixer~\cite{chord_mixer} and Cycle-MLP~\cite{cycle_mlp} are designed to process input sequences of different lengths. 
Chord-Mixer requires $\log_2 N$ layers to perform full instance interactions, where $N$ is the number of tokens.
Cycle-MLP~\cite{cycle_mlp} can be considered a combination of two 1D deformable convolutions. Therefore, the ECL of these two methods increased slowly as the number layer increases.
Compared to Chord-Mixer, SAC-MIL further divides the WSI sequence into multiple regions and performs channel shifts within each region in parallel, and also performs one extra channel shift operation to perform instance correlations.

\section{Methods}

We build SAC-MIL to encode positional information and perform full instance correlations for MLP-based methods.
The overall architecture of SAC-MIL is illustrated in Figure~\ref{fig:fig2}.
We first adopt FPS and KNN to split instances into multiple regions (Sec.~\ref{sec:method_preprocess}).
The SAC-MIL consists of an MLP layer to reduce the input feature dimension, followed by positional embedding through PROPE (Sec.~\ref{sec:method_rope}) to encode the spatial coordinates of each instance.
SAC-MIL comprises three SAC blocks (Sec.~\ref{sec:methods_SB}) to perform full instance correlations.
The final prediction is obtained by averaging all instances and using an MLP layer to predict the class.
SAC-MIL enables MLP-based methods to have the advantages of Transformer-based methods while maintaining linear computational complexity with respect to the length of the input sequence and ease of deployment.

\subsection{Preliminaries of MIL}
\label{sec:method_bg}
Given a slide $X_i$, WSI classification aims to predict the label of each slide $Y_i$.
Each slide $X_i$ is cropped into $n$ patches and processed by a feature encoder, resulting $n$ instance features \{$x_{i, 1}, x_{i, 2}, ..., x_{i, n}$ \}, where $x_{i, n} \in R^{d}$, $d$ is the dimension of each instance.
Under the MIL setting, the label of each $x_{i, j}$ is unknown during training.
A MIL classification problem can be defined as:
\begin{equation}
Y_i = 
    \begin{cases}
        0,  & \text{iff } \Sigma y_{i,j} = 1, \\
        1, & \text{otherwise}.
    \end{cases}
\end{equation}
where $y_{i,j}$ is the label for $x_{i, j}$, where $y_{i, j} \in {0, 1}$.
MIL considers each slide as a bag, and the patch features $\{y_{i, 1}, y_{i, 2}, ... y_{i, n} \}$ as the instances within each bag.


 \subsection{Divide Instances into Multiple Regions}
 \label{sec:method_preprocess}

We adopt FPS and KNN to divide instances into multiple regions in order to encode the local context.
We keep the corresponding 2D coordinates ($c^{(x)}_{i, n}$, $c^{(y)}_{i, n}$) (top-left corner of a patch) for each instance $x_{i, n}$. 
We use FPS~\cite{FPS} to select $k$ coordinates as the cluster centers and then adopt KNN to find the top $k-1$ closest instances. 
his ensures that instances within the same region are close to each other.

\noindent\textbf{Find cluster centers.}
In order to encode the local context, we evenly divide the irregular shape of each tissue area into multiple local regions, which may also have irregular shapes.
We adopt the FPS~\cite{FPS} algorithm to select $R$ cluster centers from the set 
$ C_i \in \{(c^{(x)}_{i, 1}, c^{(y)}_{i, 1}), (c^{(x)}_{i, 2}, c^{(y)}_{i, 2}), \ldots, (c^{(x)}_{i, n}, c^{(y)}_{i, n})\} $:
\begin{equation}
    j = \arg\max_{c_{i, p} \in C_i^P \setminus C_i^S} \left( \min_{c_{i, s} \in C_i^S} d(c_{i, p}, c_{i, s}) \right)
    \label{eq:fqs}
\end{equation}
where $j$ is the index of the coordinate $(c^{(x)}_{i, j}, c^{(y)}_{i, j})$ in $C_i$, $R$ is obtained by $\left\lceil k / n \right\rceil$, and $k$ is the number of instances for each region. 
$C_i^S$ is the set of selected coordinates calculated by Eq.~\ref{eq:fqs}, and $C_{i, s}$ is the $s$-th coordinate in $C_i^S$. $C_i^P$ is the remaining coordinates, and $C_{i, p}$ is the $p$-th coordinate in $C_i^P$. $d(\cdot)$ is the 2D Euclidean distance.


\begin{figure}[t]
\vspace{-1.6em}
    \centering
    \includegraphics[width=1\columnwidth]{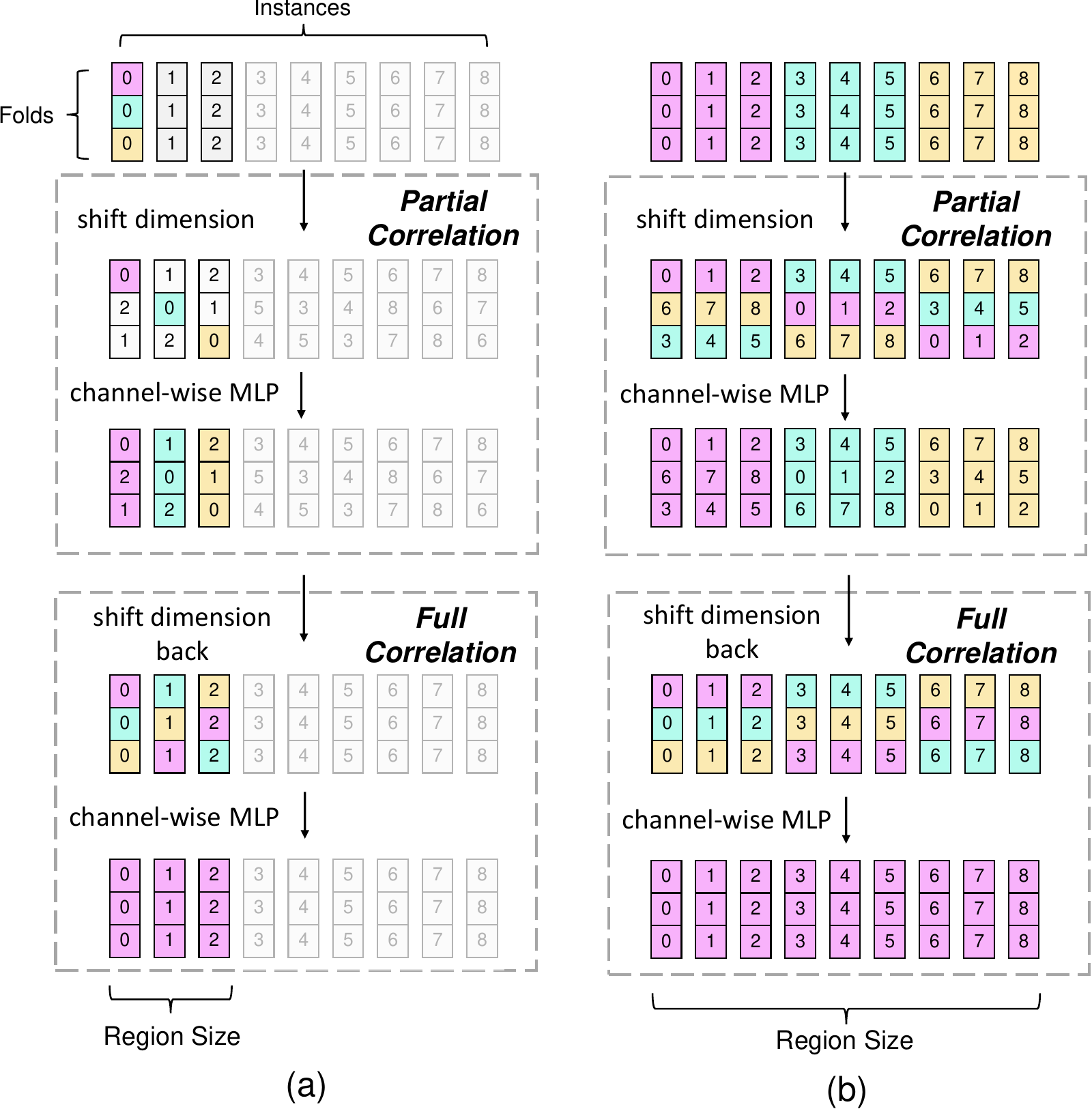}
    \vspace{-1.8em}
    \caption{
        Illustration of two stacked SAC blocks.
        (a) The first SAC block.
        (b) The second SAC block.
        Partial correlation makes each instance encode partial information of other instances, which can be seen as fold-level correlation.
        Full correlation enables each instance to encode full information of other instances; each fold contains all the information from other folds within the same region.
        As the number of SAC blocks increases, the number of fully correlated instances is exponentially increased.
        Folds with the same color indicate that they share the same information after correlation.
        If folds of more than two colors are correlated, we pick one of these colors to indicate that these folds contain the same information for simplicity.
    }
    \label{fig:blocks}
    \vspace{-0.9em}
\end{figure}

\noindent\textbf{Locate nearby instances}.
For a given coordinate $(c^{(x)}_{i, j}, c^{(y)}_{i, j})$ in $C_i^S$, we calculate the top-$(k-1)$ nearest instances to $(c^{(x)}_{i, j}, c^{(y)}_{i, j})$ using KNN.
Then, we obtain $R$ sets of coordinates, each containing $k$ instances.

\noindent\textbf{Arrange the instances}.
After obtaining $R$ sets of coordinates for a WSI sequence, we arrange the instances of the WSI such that instances within the same region are continuously placed in the WSI sequence.

After the above operations, SAC blocks will be able to encode the local context if we set the initial region size to $ k $, as discussed in Sec.~\ref{sec:methods_SB}.

\subsection{Polar Rotary Position Embedding}
\label{sec:method_rope}
We propose PROPE to encode the spatial information of instances.
PROPE normalizes the 2D coordinates $(c^{(x)}_{i, j}, c^{(y)}_{i, j})$ of each instance into the same range and converts the Cartesian coordinates into polar coordinates.
Finally, we perform positional encoding based on the normalized polar coordinates.
We briefly describe ROPE and 2D ROPE and then introduce PROPE.


\noindent\textbf{ROPE and 2D ROPE.}
ROPE~\cite{rope} is a recent method designed for NLP tasks. 
Given an input sequence $ h \in \mathbb{R}^{L \times D} $, ROPE rotates the $ m $-th instance $ h_m $ as follows:
\begin{equation}
    \hat{h}_m = h_m e^{im\theta}
\end{equation}
where $ m \in [0, L-1] $ is the token index and $ \theta $ is defined by:
\begin{equation}
    \theta_t = 10000^{-t/(D/2)}, \text{ where } t \in \{0, 1, \ldots, D/2\}.
\end{equation}
Here, $ D $ is the feature dimension.
For vision tasks, 2D ROPE~\cite{2d_rope_fit, 2d_rope_unified} is proposed to encode 2D spatial information:
\begin{equation}
    \mathbf{h}(m, 2t) = e^{i\theta_t p_m^x}, \quad \mathbf{h}(m, 2t+1) = e^{i\theta_t p_m^y}
    \label{eq:2d_rope}
\end{equation}
where $ p_m^x $ is the $ x $-axis coordinate of the reshaped input sequence, and $ p_m^y $ is the $ y $-axis coordinate of the reshaped input sequence. 

Although 2D ROPE is adopted in many vision tasks, it only encodes one axial positional information for each dimension, which can lead to incomplete 2D positional information. Additionally, ROPE-based methods suffer from the length extrapolation issue~\cite{PI}, where training and testing sequences have different lengths.

\noindent\textbf{PROPE.}
We normalize the 2D coordinates of the training and testing instances to alleviate the length extrapolation issue~\cite{PI}.

We normalize each 2D coordinate $(c^{(x)}_{i, j}, c^{(y)}_{i, j})$ as:
\begin{equation}
    \hat{c}^{(x)}_{i, j} = \frac{c^{(x)}_{i, j} - c^{(x)}_{i, \min}}{c^{(x)}_{i, \max} - c^{(x)}_{i, \min}}
\end{equation}
where $c^{(x)}_{i, \max}$ is the largest $x$-coordinate in the set $C_i$, and $c^{(x)}_{i, \min}$ is the smallest $x$-coordinate in the set $C_i$.
We perform the same normalization on $c^{(y)}_{i, j}$.
In this way, the 2D coordinates are normalized to $(0, 1)$ by dividing the difference between the maximum coordinate and the minimum coordinate of each slide.
Therefore, the radius of the polar coordinate will also be normalized to the range $(0, 1)$ when converting 2D coordinates ($c^{(x)}_{i, \max}$, $c^{(y)}_{i, \max}$) into polar coordinates.

We convert the 2D coordinates into polar coordinates to encode the complete positional information for each dimension, different from 2D ROPE~\cite{2d_rope_fit, 2d_rope_unified}.
For each coordinate $(\bar{c}^{(x)}_{i, j}, \bar{c}^{(y)}_{i, j})$, we convert 2D Cartesian coordinates into polar coordinates by:
\begin{equation}
\begin{aligned}
    \rho_{i, j} &= \sqrt{(c^{(x)}_{i, j})^2 + (c^{(y)}_{i, j})^2} \cdot \lambda \\
    \alpha_{i, j} &= \arctan2(c^{(y)}_{i, j}, c^{(x)}_{i, j})
\end{aligned}
\end{equation}
where $\rho_{i, j}$ is the radial distance (radius) and $\alpha_{i, j}$ is the rotation angle in the polar coordinate system for $(\bar{c}^{(x)}_{i, j}, \bar{c}^{(y)}_{i, j})$.
And $\lambda \in \mathbb{R}$ is the scaling factor that controls the range of $\rho_{i, j}$, ensuring that two nearby instances are more spread out in terms of their $\rho$ values~\cite{PI}.

We compute the positional information as:
\begin{equation}
    \hat{h}_m = h_m e^{i(\rho_{i, m}\theta + \alpha_{i, m})}
\end{equation}
Different from ROPE, we use the radius $\rho_{i, m}$ of each instance to replace the token index and add $\alpha_{i, m}$ into the rotation angle to further differentiate instances with the same radius.
We only adopt PROPE at the first layer of the SAC block.

RPOPE is a parameter-free positional encoding method which can handle the length exploitation issue~\cite{PI} and encodes the complete spatial information for each dimension.

\subsection{SAC Block}
\label{sec:methods_SB}

We introduce the SAC block to perform full instance correlations of sequences with various lengths.
Mixer block first applies a Layer Norm to normalize the input sequence, then divides the WSI sequence into $R$ regions, each with $k$ instances, where the initial value of $R$ is obtained by $\left\lceil \frac{L}{k} \right\rceil$.
For all instances within the same region, we perform two operations: partial correlation and full correlation to encode the full instance correlations.

\noindent\textbf{Partial Correlation.}
For a given input sequence $\hat{h} \in \mathbb{R}^{L \times D}$, we split each instance along the channel dimension into $\frac{D}{k}$ folds and shift each fold with an increased step:
\begin{equation}
    \hat{h}_{i^s, t_j} = \hat{h}_{i, t_j}
    \label{eq:shift}
\end{equation}
where
\begin{equation}
    i^s = \left( i + \left\lfloor \frac{t_j}{k} \right\rfloor \cdot k^l \right) \mod k^{l+1}
    \label{eq:mod}
\end{equation}
For the $t_j$-th dimension of the $i$-th instance $\hat{h}_{i}$, we shift it to the same dimension of the $i^s$-th instance.
The token index $i^s$ is determined by the shift factor $k^l$, fold index $\left\lfloor \frac{t_j}{k} \right\rfloor$, and the region size $k^{l+1}$. 
Here, $l \in [0, 1, ..., N]$ represents the layer index of SAC blocks, where $N$ is the number of SAC blocks.
For the $\left\lfloor \frac{t_j}{k} \right\rfloor$-th fold of each instance, it shifts $(i + \left\lfloor \frac{t_j}{k} \right\rfloor \cdot k^l)$ steps. 
Note that we use $\mod$ the region size $k^{l+1}$ in Eq.~\ref{eq:mod}, because the SAC block computes the shifted position in parallel over multiple regions.
Therefore, this shift operation is cheap to implement using CUDA. 
It can also be done in PyTorch with the roll function.
The pseudo code of dimension shift operation is illustrated in Appendix~\ref{se:pseudo-code}.


%
The SAC block adopts a channel-wise MLP layer to perform instance correlations:
\begin{equation}
    \hat{h}_{i^{o1}} = W_1 \hat{h}_{i^s}
\end{equation}
where $W_1$ is the channel-wise MLP layer and $\hat{h}_{i^{o1}}$ is the output of $W_1$. 
The SAC block uses a channel-wise MLP to process input sequences of various lengths. 
At this step, the SAC block essentially performs fold-level correlations, as each instance contains folds from other instances within the same region. 
Therefore, each fold only encodes partial information within the region.

\begin{table*}[!t]
\vspace{-1.6em}
\centering
\resizebox{\linewidth}{!}{
\begin{tabular}{c | c  c  c  | c  c  c  | c c c } 
\hline
  \multirow{2}{*}{Methods}         & \multicolumn{3}{c|}{CAMELYON-16}                                          & \multicolumn{3}{c|}{TCGA-LUNG}           &  \multicolumn{3}{c}{TCGA-BRAC}   \\ 
                                   & ACC & AUC &  F1
                                   & ACC & AUC &  F1
                                   & ACC & AUC &  F1
                                   \\ 
  \hline

  ab-mil~\cite{abmil}  &    $83.5_{1.5}$      &  $84.3_{1.9}$          & $79.1_{1.7}$ 
  &  $87.5_{0.5}$	       & $91.4_{1.4}$	 & $87.5_{0.5}$ 
  &  $83.3_{3.8}$        & $86.6_{5.0}$                & $80.3_{3.2}$
  \\
  ds-mil~\cite{dsmil}  & $85.6_{3.6}$             &  $87.3_{2.2}$          &
  $80.0_{2.0}$               &
  $88.9_{1.9}$             & $94.3_{6.0}$     & $88.1_{1.9}$  
  &  $84.1_{4.9}$	 & $89.4_{3.8}$	 &  $82.7_{3.6}$\\
  CLAM-SB~\cite{clam}  & $83.5_{3.3}$  &  $86.3_{3.3}$	   & $80.3_{1.1}$                &  $86.3_{3.4}$	         &$91.3_{2.1}$	               & $85.4_{1.3}$  & $84.0_{3.3}$	                  &   $87.1_{2.4}$	     &$81.5_{6.8}$	               \\
  CLAM-MB~\cite{clam}  & $83.7_{3.3}$  &  $86.4_{3.3}$	   & $80.5_{2.2}$    
  &  $88.3_{3.4}$	         &$92.9_{2.1}$	               & $88.2_{1.0}$  &
  $82.7_{2.3}$	                  &   $86.8_{4.0}$	     &$80.5_{4.6}$	               \\
  DTFD-MIL~\cite{dtfd}  & $87.1_{0.6}$ & $90.4_{1.2}$	       & $86.5_{1.3}$	            
  &  $88.3_{1.4}$	        & $93.2_{2.3}$	            & $88.8_{1.5}$	              
  &   $84.4_{4.6}$	        &$87.1_{1.6}$	               & $84.3_{3.2}$\\
  IBMIL-MIL (DTFD)~\cite{ibmil}  & $87.4_{0.4}$ & $89.5_{1.2}$	    & $87.0_{1.3}$
    &  $89.4_{2.7}$	         &  $93.7_{1.1}$            & $89.3_{0.5}$     
    &  $84.8_{3.3}$	         & $87.9_{1.4}$	            & $83.2_{1.1}$\\
  \hline
  Mamba-MIL~\cite{mambamil}   & $86.6_{0.6}$ & $94.7_{1.2}$	       & $85.8_{1.3}$	            &  $87.8_{2.7}$	         & $92.9_{1.6}$	            & $87.7_{2.8}$	              
  &$83.3_{3.4}$      & $86.5_{1.0}$              & $83.2_{1.6}$\\
    Mam-MIL~\cite{mammil}   & $88.6_{0.6}$ & $94.9_{1.2}$	       & $87.8_{1.3}$	            &  
     $87.2_{2.7}$	         & $91.9_{1.6}$	            & $87.1_{2.8}$	            
    & $82.3_{4.0}$      & $85.5_{1.2}$              & $82.2_{3.6}$\\
  Trans-MIL~\cite{transmil} 
  & $86.0_{4.4}$	     &  $90.1_{6.2}$	  &  $85.2_{1.2}$	            &  $88.5_{2.1}$	         & $94.3_{0.8}$	      &$88.4_{0.1}$ 	   
  &  $83.4_{2.0}$	      & $88.7_{5.2}$	      &$79.7_{3.4}$ \\
  HIPT~\cite{hipt} & $80.8_{1.3}$ & 	$85.1_{3.7}$   &  $79.4_{3.5}$        
  &   $87.3_{2.4}$	         &	$94.4_{2.2}$                & $86.6_{2.7}$	          
  &   $84.8_{4.2}$	           &$86.1_{2.7}$	                & $84.5_{1.3}$\\
  R$^{2}$T-MIL~\cite{re_embedding} & $87.6_{0.3}$ & $92.4_{1.2}$	       & $88.1_{1.1}$	            &  $88.3_{1.2}$	         & $93.1_{1.6}$	            & $87.7_{2.8}$	     
  &  $84.3_{4.0}$      & $87.1_{1.7}$              & $84.2_{0.4}$\\
  Norma~\cite{norma}  & $89.6_{0.6}$ & $95.1_{1.2}$	       & $88.8_{1.3}$	            &  $89.8_{2.7}$	         & $93.9_{1.6}$	            & $89.3_{2.8}$	    
  &  $84.8_{4.0}$      & $88.5_{1.2}$              & $83.2_{3.6}$\\
  \hline
  SAC-MIL (Ours)   & \bm{$90.3_{0.6}$}   & \bm{$96.1_{1.2}$}	       & \bm{$89.3_{1.6}$}	              
  & \bm{$90.4_{0.6}$}	  & \bm{$95.0_{1.4}$}       & \bm{$90.1_{0.8}$}	             
  &  \bm{$85.4_{4.2}$}      & \bm{$89.1_{0.5}$}              & \bm{$84.9_{1.3}$}\\
  \hline
\end{tabular}
}
\vspace{-0.9em}
\caption
    {
         The performance of MLP-based and Transformer-based methods is evaluated on the CAMELYON-16, TCGA-LUNG, and TCGA-BRAC datasets with ResNet50 feature.
    }
\label{table:t1}
\vspace{-0.9em}
\end{table*}

\noindent\textbf{Full Correlation.}
For a given sequence $\hat{h}_i^{o1} \in \mathbb{R}^{L \times D}$, we shift each fold to its original position and adopt another channel-wise MLP to perform instance correlation.
The shift operation is the same as Eq.~\ref{eq:shift}, but with a different direction:
\begin{equation}
    \hat{h}_{i^{{o1\_s}}, t_j} = \hat{h}_{i^{o1}, t_j}
    \label{eq:shift2}
\end{equation}
where
\begin{equation}
    i^{o1\_s} = \left( i^{o1} - \left\lfloor \frac{t_j}{k} \right\rfloor \cdot k^l \right) \mod k^{l+1}
    \label{eq:mod2}
\end{equation}
Eq.~\ref{eq:shift2} and Eq.~\ref{eq:mod2} shift each fold to its original position.

SAC-MIL adopts another channel-wise MLP to perform instance correlations:
\begin{equation}
    \hat{h}_{i^{o2}} = W_2 \hat{h}_{i^{o1\_s}}
\end{equation}
where $W_2$ is the MLP layer and $\hat{h}_{i^{o2}}$ is the output instance from $W_2$.
At this step, since each fold contains information from other folds in different instances, $W_2$ aggregates information stored in other folds within the same instance.
All the information of all folds within each region will be aggregated through $W_2$.
Therefore, $W_2$ performs full instance correlations within each region.

After partial correlation and full correlation steps, each fold contains all the information of other folds within the same region.
The region size will exponentially increase as $l$ increases, illustrated in Eq.~\ref{eq:mod}.
We perform the same operations to aggregate information across regions, as illustrated in Fig.~\ref{fig:blocks}.
By adopting this approach, we can gradually perform full instance correlation by exponentially increasing the region size.

\begin{table*}[t]
\vspace{-2.0em}
\centering
\resizebox{\linewidth}{!}{
\begin{tabular}{c | c  c  c  | c  c  c  | c c c } 
\hline
  \multirow{2}{*}{Methods}         & \multicolumn{3}{c|}{CAMELYON-16}                                          & \multicolumn{3}{c|}{TCGA-LUNG}           &  \multicolumn{3}{c}{TCGA-BRAC}   \\ 
                                   & ACC & AUC &  F1 &  ACC  & AUC & F1 &  ACC  & AUC &  F1 \\ 
  \hline

  AB-MIL~\cite{abmil}  
  &    $90.7_{1.3}$      &  $91.3_{1.3}$          & $90.1_{0.3}$    
  &  $95.1_{1.1}$             & $97.7_{0.5}$                & $95.5_{1.0}$                   &   
  $91.9_{3.0}$	       & $95.7_{2.1}$	 & $89.3_{1.0}$ \\
  DS-MIL~\cite{dsmil}  
  & $91.6_{1.3}$              & $94.7_{3.3}$                & $91.2_{1.6}$ 
  & $94.4_{1.1}$             & $97.9_{0.5}$                & $94.4_{1.0}$                   
  &  $89.1_{3.1}$	 & $95.6_{2.6}$	 &  $87.0_{1.4}$\\
  CLAM-SB~\cite{clam}  
  &  $93.5_{1.3}$    &  $95.3_{0.3}$	   & $91.3_{1.1}$                
  &  $95.2_{0.8}$	   &  $98.0_{0.4}$	   & $95.0_{0.8}$  
  & $90.5_{4.0}$	                  &   $95.1_{2.0}$	     &$89.2_{2.6}$	               \\
  CLAM-MB~\cite{clam}  
  &  $93.7_{3.3}$  &  $95.4_{3.3}$	   & $90.5_{2.2}$                
  &  $95.4_{0.7}$	 &  $98.3_{0.7}$	   & $95.3_{0.6}$  
  & $90.8_{3.3}$	                  &   $95.8_{2.2}$	     &$88.1_{3.7}$	               \\
 DTFD-MIL~\cite{dtfd}  
 & $93.4_{0.6}$ & $97.5_{1.2}$	       & $93.5_{1.3}$	          
 &  $94.9_{1.4}$	         & $98.2_{2.3}$	            & $94.8_{1.5}$	           
  &   $91.4_{4.6}$	      &$96.0_{1.6}$	               & $90.3_{3.2}$\\
  IBMIL-MIL (DTFD)~\cite{ibmil}  & $95.4_{0.4}$ & $89.5_{1.2}$	       & $87.0_{1.3}$	            &  $94.4_{2.7}$	         &  $98.4_{1.1}$            & 	   $94.2_{0.5}$          
  &   $91.6_{3.3}$	      & $96.4_{1.4}$	               & $90.5_{1.1}$\\
  \hline
  Mamba-MIL~\cite{mambamil}   
  & $89.6_{0.6}$ & $96.1_{1.2}$	       & $88.8_{1.3}$	            &  $92.8_{2.7}$	         & $96.9_{1.6}$	            & $92.7_{2.8}$	             
  &  $91.3_{4.0}$      & $95.5_{1.2}$              & $89.2_{3.6}$\\
    Mam-MIL~\cite{mammil}   & $91.1_{0.6}$ & $96.4_{1.2}$	       & $90.8_{1.3}$	            &  $93.8_{2.7}$	         & $96.9_{1.6}$	            & $93.4_{1.8}$	             
    &  $91.4_{1.3}$      & $95.5_{1.2}$              & $89.2_{3.6}$\\
  Trans-MIL~\cite{transmil} 
   & $92.7_{4.4}$	     &  $98.1_{1.2}$	  &  $92.1_{2.6}$	  
  &  $95.2_{1.1}$	         & $98.3_{1.0}$	            &  $95.1_{1.0}$ 	         
  &  $91.6_{4.2}$	      & $95.6_{2.4}$	        &   $89.1_{3.2}$ \\
  HIPT~\cite{hipt} & $90.8_{1.3}$ & 	$95.8_{3.7}$   &  $89.4_{3.5}$                
  &   $94.3_{2.4}$	         &	$95.4_{2.2}$                & $94.1_{2.7}$	                  
  &   $90.8_{4.2}$	           &$94.1_{2.7}$	                & $90.5_{1.3}$\\
  R$^{2}$T-MIL~\cite{re_embedding} & $92.6_{0.3}$ & $97.6_{1.2}$	       & $91.8_{1.1}$	          
  &  $94.9_{1.2}$	         & $95.1_{1.6}$	            & $93.7_{2.8}$	              
  &  $90.3_{4.0}$      & $95.1_{1.7}$              & $89.2_{0.4}$\\
  Norma~\cite{norma} 
  & $93.4_{0.6}$ & $96.1_{1.2}$	       & $91.1_{1.3}$	            
  &  $95.8_{2.7}$	         & $97.9_{1.6}$	            & $93.7_{2.8}$	             
  &  $91.3_{4.0}$      & $95.5_{1.2}$              & $91.2_{3.6}$\\
  \hline
  SAC-MIL   & \bm{$94.1_{0.6}$}   & \bm{$98.5_{1.2}$}	       & \bm{$92.5_{1.6}$}	              & \bm{$96.4_{1.1}$}	         & \bm{$98.7_{1.1}$}	            & \bm{$95.7_{1.0}$}	              &
  \bm{$92.1_{2.8}$}      & \bm{$96.7_{1.8}$}              & \bm{$91.5_{3.0}$}\\
  \hline
\end{tabular}
}
\vspace{-0.9em}
\caption
    {
         The performance of MLP-based and Transformer-based methods is evaluated on the CAMELYON-16, TCGA-LUNG, and TCGA-BRAC datasets with the UNI~\cite{uni} feature.
    }
\label{table:t2}
\vspace{-0.9em}
\end{table*}

\section{Experiment}
We conducted a series of experiments to evaluate the effectiveness of the proposed SAC-MIL.
The experimental setups are detailed below:

\noindent\textbf{Training:} 
We use the Adam optimizer during training. 
The learning rate is set to $1 \times 10^{-4}$. 
The batch size is set to 1, and the number of training epochs is set to 200.
The experiments are conducted using an NVIDIA 3090 GPU.

\noindent\textbf{Datasets:} 
We evaluate SAC-MIL using three public datasets: CAMELONY16~\cite{cam16}, The Caner Genome Atlas (TCGA) Lung Adenocarcinoma (LUNG), TCGA Breast Invasive Carcinoma (BRRA).
Please refer to the Appendix~\ref{se:a_datasets} for more details.



\noindent\textbf{Model:}
The number of blocks $N$ is set to 3.
The initial region size $k$ is set to 64, other two region sizes are 4096, 262114 resepectively.
The model dimension $D$ is set to 512.
And we also empirically set the scale factor $\lambda$ in ROPE to 512.
We adopt ResNet50 feature if not specified.

\noindent\textbf{Evaluation:}
We assessed the performance of SAC-MIL using three metrics: F1-score, Area Under the Curve (AUC), and slide-level Accuracy (ACC).
To ensure robust evaluation, we split the training slides of CAMELYON16 and all slides from TCGA-LUNG and TCGA-BRCA into five folds. 
We then performed five-fold cross-validation on these splits to assess model performance.
We report both the mean and the standard deviations of these metrics across the five-fold validation.

\subsection{Comparison to State-of-the-Art Methods}

We compare our model with two categories of methods: Transformer-based and MLP-based methods, on three different datasets: CAMELYON-16, TCGA-Lung, and TCGA-BRAC. The Transformer-based methods include TransMIL~\cite{transmil}, Norma~\cite{norma}, R$^{2}$T-MIL~\cite{re_embedding}, HIPT~\cite{hipt}. Additionally, we also conduct experiments using two mamba-based methods: Mam-MIL~\cite{mammil}, Mamba-MIL~\cite{mambamil}. The MIL-based methods include AB-MIL~\cite{abmil}, DS-MIL~\cite{dsmil}, CLAM-SB~\cite{clam}, CLAM-MB~\cite{clam}, DTFD-MIL~\cite{dtfd}, IBMIL-MIL (DTFD)~\cite{ibmil}.

We use two different patch feature extractors: ResNet50 and UNI model. As shown in Table~\ref{table:t1}, SAC-MIL achieves 1\% performance gains over the second-best methods in AUC metric on the CAMELYON-16 dataset. 
On the TCGA-Lung dataset, SAC-MIL shows improvements of 0.7\% in AUC compared to the second-best method. 
For the TCGA-BRAC dataset, SAC-MIL also achieved the best performance.
As illustrated in Table~\ref{table:t2}, when using the UNI model as the patch feature extractor, SAC-MIL achieves 0.6\%, 0.4\% and 0.4\% performance gains over the second-best methods in ACC, AUC, and F1 metrics on the CAMELYON-16 dataset. 
On the TCGA-Lung dataset, SAC-MIL improves by 0.6\%, 0.3\% and 0.4\% in ACC, AUC, and F1 compared to the second-best method. 
For the TCGA-BRAC dataset, SAC-MIL also achieved the best performance.
The UNI model demonstrates superior results compared to the ResNet50 model, indicating that a better foundation model can enhance overall performance.

\begin{table}[t]
\centering
\resizebox{\linewidth}{!}{
\begin{tabular}{c  c c c } 
\toprule[\heavyrulewidth]
Methods & Accuracy & AUC & F1 \\ 
\hline
Sinusoidal~\cite{transformer}  
& $87.7_{3.3}$           &$94.5_{1.7}$	   & $87.5_{1.2}$                \\
Alibi~\cite{alibi}  
&  $88.5_{1.0}$        & $95.3_{1.9}$            & $88.1_{1.7}$                 \\
ROPE~\cite{rope}  
& $89.1_{3.6}$             & $94.9_{2.2}$          & $88.5_{4.3}$                 \\
2D ROPE~\cite{2d_rope_unified}  
&  $89.6_{3.6}$             & $95.3_{2.2}$           & $88.6_{4.3}$                 \\
PEG~\cite{re_embedding}  
& $89.4_{3.2}$           &$94.3_{3.3}$	     & $87.9_{1.1}$                 \\
PPEG~\cite{transmil}  
& $89.7_{3.3}$      & $96.4_{0.8}$	   & $89.1_{0.9}$                 \\
\hline
w/o PROPE   
& $87.8_{0.9}$   & $94.3_{1.1}$	       & $87.8_{1.3}$	              \\
PROPE   
& \bm{$90.3_{0.6}$}   & \bm{$96.1_{1.2}$}	       & \bm{$89.3_{1.6}$}	             \\
\bottomrule[\heavyrulewidth]
\end{tabular}
}
\vspace{-0.8em}
\caption
    {
         The comparison results of various positional encoding methods on the CAMELYON-16 dataset.
    }
\label{table:t3}
\vspace{-0.4em}
\end{table}

\subsection{Ablation Study}
\noindent\textbf{Effects of PROPE.}
Tab.~\ref{table:t3} presents the performances of different positional encoding approaches on the CAMELYON-16 dataset.
PROPE achieved the best performance compared to other positional encoding methods. Compared to ROPE, our method increased the AUC score by $1.2\%$ on the CAMELYON-16 dataset.
We attribute this improvement to the normalized polar coordinates, which can encode spatial information of each instance.
Although sinusoidal positional encoding is also a parameter-free method, we do not observe a notable improvement in the AUC score. This might be due to the length extrapolation issue.
Additionally, PROPE improved the AUC metric by $0.7\%$ for the TCGA-LUNG dataset and by $1.3\%$ for the TCGA-BRAC dataset.
PROPE has achieved higher performance compared to other positional encoding methods, demonstrating the effectiveness of our approach. 
The scaling factor $\lambda$ of PROPE is empirically set to 512, as illustrated in Appendix~\ref{se:scaling}
We additionally report comparison results on the TCGA-LUNG and TCGA-BRAC datasets in Appendix~\ref{se:a_PROPE_}.

\begin{figure}[t]
    \centering
    \vspace{-1.7em}
    \includegraphics[width=0.75\linewidth]{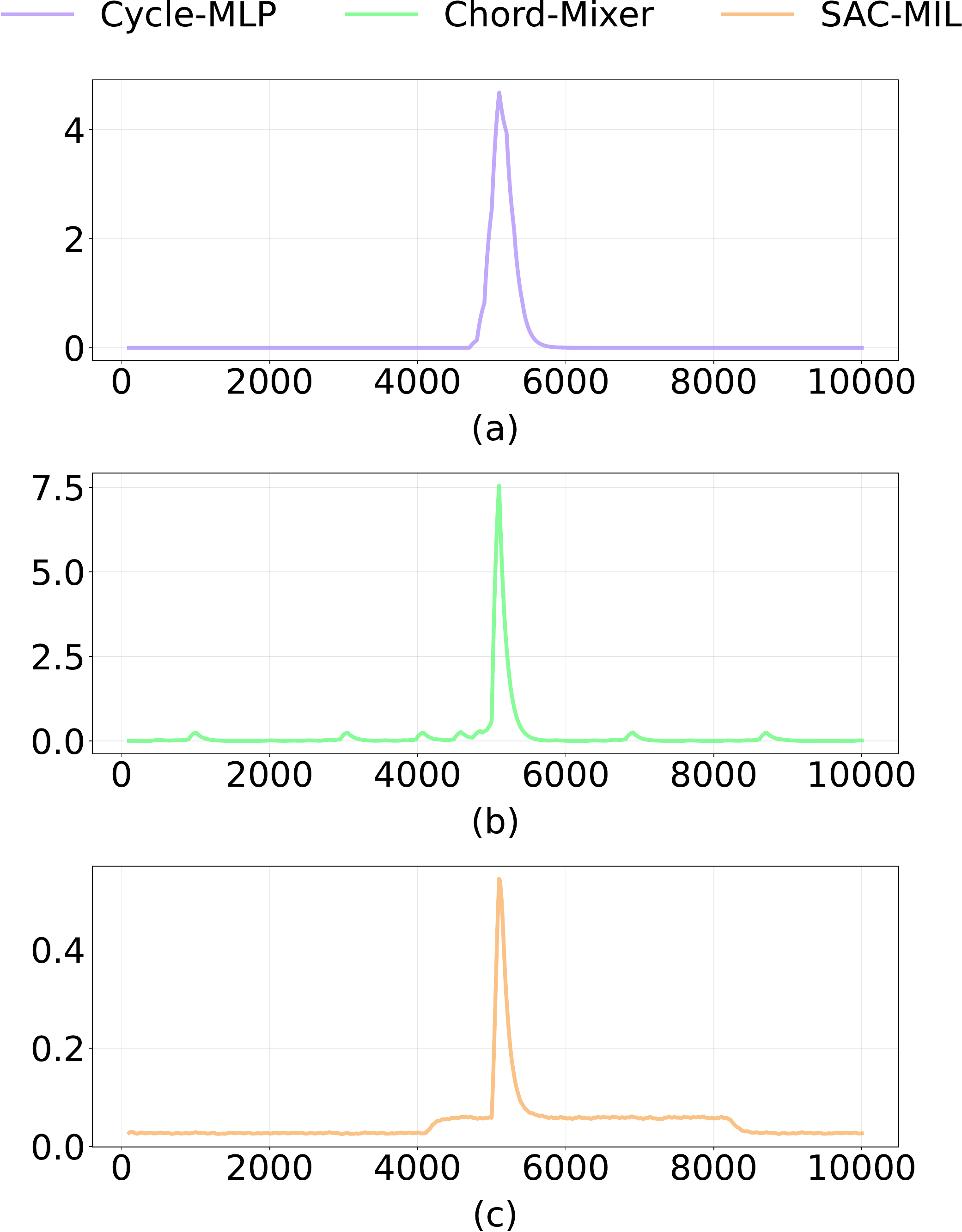}
    \vspace{-0.9em}
    \caption{
    The L2 norm of the differences between the outputs from two forward passes is calculated. 
    The input sequence length is set to 10,000.
    The x-axis represents the instance index, and the y-axis represents the L2 norm value. 
    (a) Cycle-MLP, (b) Chord-Mixer, and (c) SAC-MIL.
        }
    \label{fig:ex_diff_value}
    \vspace{-0.9em}
\end{figure}

\begin{figure}[t]
    \centering
    \vspace{-1.4em}
    \includegraphics[width=0.7\linewidth]{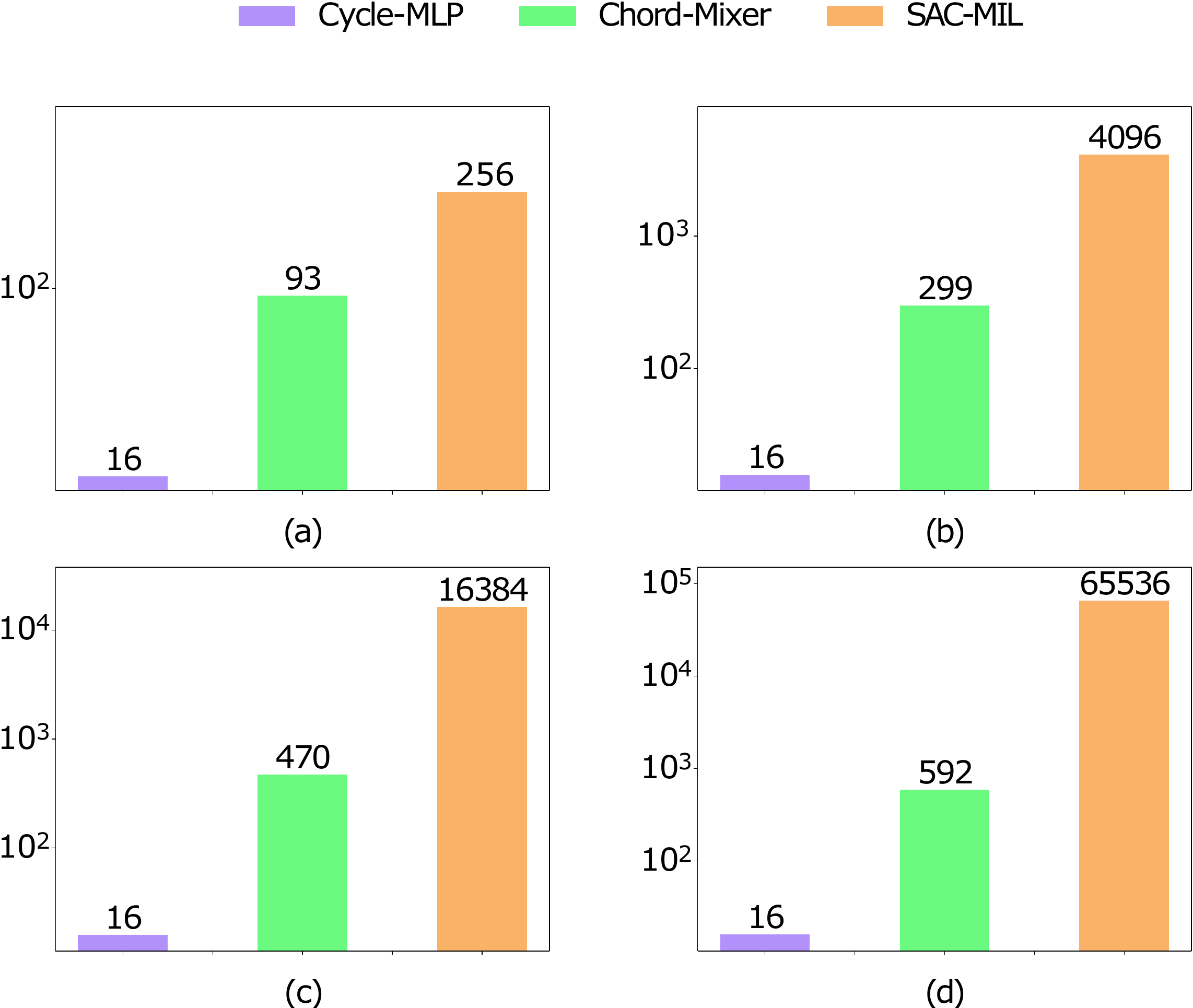}
    \vspace{-0.9em}
    \caption{
        Number of instances changes for three methods after setting the middle instance to a zero vector.
        (a), (b), (c), and (d) represent the WSI sequence lengths of 512, 4096, 16384, and 65536, respectively.
        }
    \label{fig:ex_num_diffs}
    \vspace{-0.6em}
\end{figure}

\begin{figure}[b]
\vspace{-1.0em}
    \centering
    \includegraphics[width=0.8\linewidth]{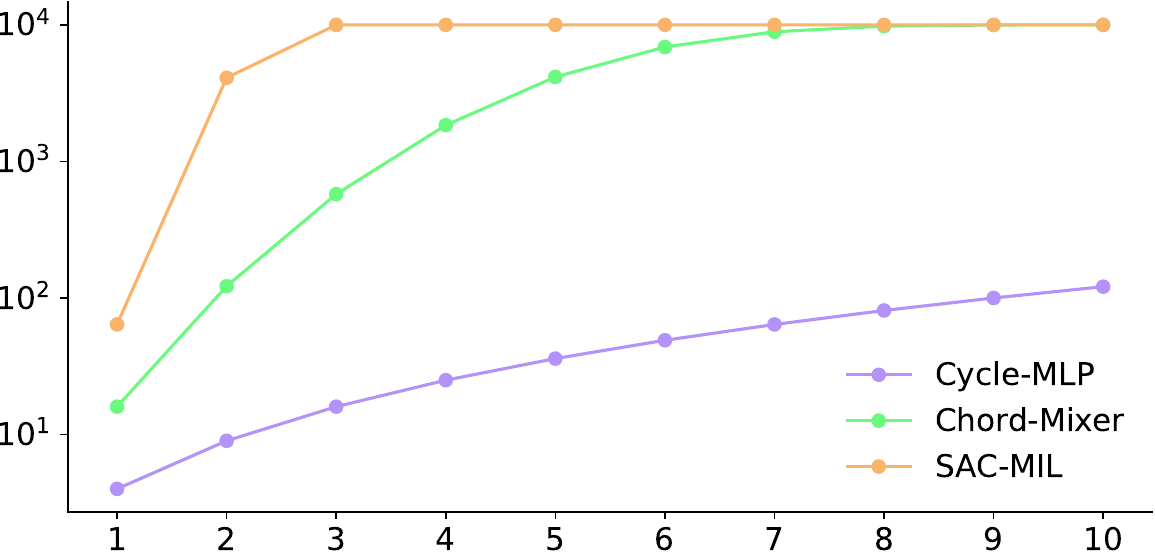}
    \vspace{-0.6em}
    \caption{
        The number of affected instances as the number of layers increases for three methods.
        The length of the input sequence is set to 10,000.
        The x-axis represents the number of layers, and the y-axis represents the number of affected instances.
        The three methods evaluated are Cycle-MLP, Chord-Mixer, and SAC-MIL.
        }
    \label{fig:layer_increase}
    \vspace{-0.9em}
\end{figure}

\begin{table}[t]
\centering
\resizebox{\linewidth}{!}{
\begin{tabular}{c  c c c } 
\toprule[\heavyrulewidth]
Methods & Accuracy & AUC & F1 \\ 
\hline
  Cycle-MLP~\cite{cycle_mlp}  
  & $89.5_{1.5}$              & $92.5_{3.5}$                & $89.0_{1.8}$  \\
  Chord-Mixer~\cite{chord_mixer}  
  &  $90.0_{1.4}$    &  $93.0_{0.5}$	   & $89.5_{1.2}$                 \\
  Nyström~\cite{Nystrom}  
  &  $88.5_{1.6}$    &  $92.0_{0.4}$	   & $88.0_{1.3}$                \\
  Flash-Attn~\cite{flash}  
  &  $89.0_{1.5}$    &  $93.5_{0.4}$	   & $89.0_{1.2}$                 \\
  \hline
  SAC-MIL  (Ours)
    & \bm{$90.3_{0.6}$}   & \bm{$96.1_{1.2}$}	       & \bm{$89.3_{1.6}$}	              \\
\bottomrule[\heavyrulewidth]
\end{tabular}
}
\vspace{-0.9em}
\caption
    {
         The comparison of various instance correlation methods on the CAMELYON-16 dataset.
    }
\label{table:blocks_cam16}
\vspace{-0.9em}
\end{table}
\noindent\textbf{Effective Context Length of SAC Block.}
We compare the ECL of SAC blocks with Cycle-MLP~\cite{cycle_mlp} and Chord-Mixer~\cite{chord_mixer} in Fig.~\ref{fig:ex_num_diffs}, as both methods can also process sequences of various lengths.
We first randomly initialize an input sequence and three models.
Each model is stacked with three layers of Cycle-MLP, Chord-Mixer, and SAC blocks, respectively.
We then perform two forward passes using the same input sequence on these three models.
The first forward pass takes the randomly initialized input sequence.
In the second forward pass, we set the middle instance ($\left\lfloor \frac{N}{2} \right\rfloor$-th) of the same input sequence used in the first forward pass to a zero vector.
We count the number of instance changes between the output sequences from the two forward passes to evaluate ECL.
For the input sequences, we adopt four different lengths: 256, 4096, 16384, and 65536.
In Fig.~\ref{fig:ex_num_diffs}, the ECL of Cycle-MLP remains at 16 for different lengths, which is expected since cyclic MLP can be seen as a stack of two 1D deformable convolutions~\cite{deformable}.
The ECL of Chord-Mixer is larger than that of the SAC block because Chord-Mixer requires $\log_2 N$ layers to perform full instance correlations~\cite{chord_mixer}.
Due to the partial and full correlation operations, the SAC block can perform full instance correlations.
In Fig.~\ref{fig:ex_diff_value}, all output instances of the SAC block change after setting the $\left\lfloor \frac{N}{2} \right\rfloor$-th instance to a zero vector, validating that the ECL of SAC is equal to the sequence length.
Fig.~\ref{fig:layer_increase} further denotes the ECL of SAC block can increase rapidly compared when the number of layer increases compared to other methods.

\noindent\textbf{Compared with other Instance Correlation Methods.}
Tab.~\ref{table:blocks_cam16} investigates the effectiveness of different instance correlation methods.
We replaced the SAC blocks within SAC-MIL with various methods and used the same hyper-parameters.
SAC-MIL increases by 2.6\% on the AUC metric compared to Flash-Attn~\cite{flash}, and Transformer-based methods do not significantly outperform MLP-Mixer based methods, indicating that under small training sample settings, Transformer-based methods may not have more advantages over MLP-style methods.
SAC-MIL outperforms other methods, demonstrating the effectiveness of SAC blocks.

\subsection{Visualization}
\begin{figure}[t]
\vspace{-1.4em}
    \centering
    \includegraphics[width=0.75\linewidth]{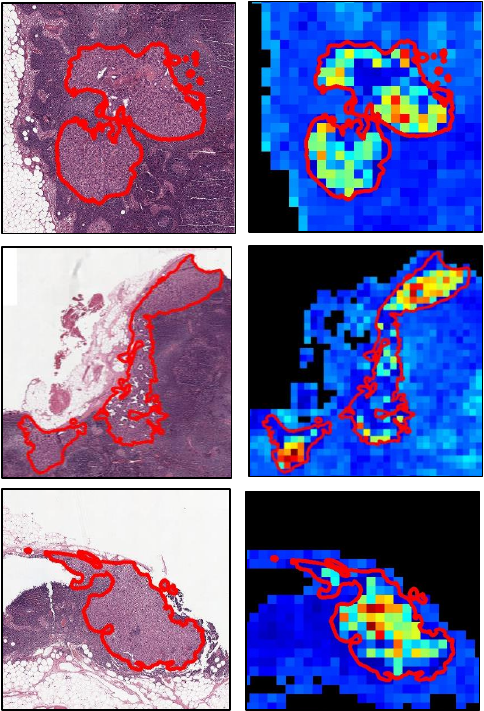}
    \vspace{-0.4em}
    \caption{
        Patch visualization produced by SAC-MIL.
        Images in the left column is the ground truth of tumor region (red).
        Left Column:Ground truth images of the tumor regions, highlighted in red.
        Right Column: The L2 norm of each output instance.
        }
    \label{fig:vis}
    \vspace{-0.9em}
\end{figure}

We visualize the predictions of SAC-MIL in Fig.~\ref{fig:vis} on the CAMELYON-16 dataset.
We calculate the L2 norm of the output from the last MLP layer of SAC-MIL.
As shown in Fig.~\ref{fig:vis}, instances within the tumor region are effectively located, with nearly no extra instances identified outside the tumor border area. This indicates that our method has a low false positive rate for precise tumor instance detection. The clear delineation of tumor boundaries highlights the robustness and accuracy of our approach.
Most of the tumor areas can be accurately localized, demonstrating the effectiveness of SAC-MIL in handling complex tissue structures.
More visualization results can be found in Appendix~\ref{sec:a_vis}.

\section{Conclusion}
We propose SAC-MIL for performing WSI classification.
SAC-MIL adopts PROPE to encode positional information.
PROPE uses polar coordinate systems to encode the complete positional information of each dimension.
It also normalizes all the polar coordinates of training and testing samples to alleviate the length exploration issue~\cite{PI}.
SAC-MIL introduces a SAC block to perform full instance correlation, which can exponentially increase the instance correlation level.
SAC block can perform full instance correlation and encode positional information with only linear computational complexity, and the simple MLP-style structure making it friendly for deployment.
We will explore the application of SAC-MIL on various WSI analysis tasks, such as survival prediction, and also aim to compare our SAC block with the Transformer block in multimodal settings in the future.

\appendices

\section{Pseudo Code of $f_s$ in SAC-MIL}
\label{se:pseudo-code}

The pseudo code of channel shift operation $f_s$ in SAC-MIL is illustrated in~\ref{alg:fs_function}.

\begin{algorithm*}[t]
\caption{Shift Function $f_s$}
\label{alg:fs_function}
\begin{algorithmic}[1]
\Function{$f_s$}{$x$}
    \If{$x.\text{shape}[-2] == 0$}
        \State \Return $x$
        \Comment{Check if the second last dimension is zero}
    \EndIf

    \State $B, L, D \gets \text{shape of } x$
    \Comment{Get the shape of tensor $x$}


    \State chunks $\gets$ split($x$, $k$, dim=-1)
    \Comment{Split $x$ along the last dimension into $k$ chunks, each with size $D // k$}

    \State res $\gets [\,]$
    \Comment{Initialize an empty list to store results}

    \For{$idx = 0$ \textbf{to} $k-1$}
        \State chunk $\gets$ chunks[$idx$]
        \Comment{Get the $idx$-th chunk from the split}
        \State shifted\_chunk $\gets$ roll(chunk, $idx \times \text{scale}$, dim=-2)
        \Comment{Shift the chunk along the second last dimension}
        \State append(res, shifted\_chunk)
        \Comment{Append the shifted chunk to the result list}
    \EndFor

    \State res $\gets$ concatenate(res, dim=-1)
    \Comment{Concatenate all chunks in res along the last dimension}

    \If{$L \geq \text{region\_size}$}
        \State res $\gets$ reshape(res, $(B, L, D)$)
        \Comment{Reshape res back to dimensions $(B, L, D)$}
    \EndIf

    \State \Return res
    \Comment{Return the final concatenated tensor}
\EndFunction
\end{algorithmic}
\caption{Pseudo code of channel shift operation}
\end{algorithm*}

\section{Datasets}
\label{se:a_datasets}
We evaluate SAC-MIL using three public datasets: CAMELONY16~\cite{cam16}, The Caner Genome Atlas (TCGA) Lung Adenocarcinoma (LUNG), TCGA Breast Invasive Carcinoma (BRRA).
The CAMELYON16 dataset consists of 270 training slides and 130 testing slides. 
The TCGA-LUNG dataset contains 875 slides, and the TCGA-BRCA dataset contains 1053 slides.
To ensure robust evaluation, we split the training slides of CAMELYON16 and all slides from TCGA-LUNG and TCGA-BRCA into five folds.

\section{Effects of PROPE.}
\label{se:a_PROPE_}

\begin{table}[!t]
\centering
\resizebox{\linewidth}{!}{
\begin{tabular}{c  c c c } 
\toprule[\heavyrulewidth]
Methods & Accuracy & AUC & F1 \\ 
\hline
Sinusoidal~\cite{transformer}  
& $89.3_{3.4}$ & $93.8_{2.1}$ & $88.6_{1.0}$ \\
Alibi~\cite{alibi}  
& $89.5_{1.9}$ & $94.7_{1.8}$ & $85.3_{2.6}$ \\
ROPE~\cite{rope}  
& $89.8_{1.6}$ & $94.4_{2.1}$ & $86.1_{1.1}$ \\
2D ROPE~\cite{2d_rope_unified}  
& $89.8_{1.6}$ & $94.4_{2.1}$ & $89.7_{1.1}$ \\
PEG~\cite{re_embedding}  
& $88.9_{3.4}$ & $94.1_{2.1}$ & $88.4_{1.3}$ \\
PPEG~\cite{transmil}  
& $89.5_{3.4}$ & $94.5_{2.1}$ & $89.6_{1.0}$ \\ \hline
w/o PROPE   
& $89.7_{1.6}$ & $94.3_{0.5}$ & $89.4_{1.8}$ \\
PROPE   
& $\mathbf{90.4_{0.6}}$ & $\mathbf{95.0_{1.4}}$ & $\mathbf{90.1_{0.8}}$ \\ 
\bottomrule[\heavyrulewidth]
\end{tabular}
}
\caption
    {
         The comparison results of various positional encoding methods on the TCGA-LUNG dataset.
         The instance feature extractor utilizes the ImageNet-pretrained ResNet50~\cite{resnet}.
         The results reported include the average and standard deviation of five runs.
         The highest performance is highlighted in bold. 
    }
\label{table:a_lung_prope}
\end{table}

\begin{table}[!t]
\centering
\resizebox{\linewidth}{!}{
\begin{tabular}{c  c c c } 
\toprule[\heavyrulewidth]
Methods & Accuracy & AUC & F1 \\ 
\hline
Sinusoidal~\cite{transformer}  
& $83.8_{0.3}$ & $87.1_{2.4}$ & $83.3_{6.8}$ \\
Alibi~\cite{alibi}  
& $84.4_{2.1}$ & $87.5_{1.9}$ & $84.5_{3.2}$ \\
ROPE~\cite{rope}  
& $84.8_{4.9}$ & $88.8_{5.4}$ & $84.0_{6.7}$ \\
2D ROPE~\cite{2d_rope_unified}  
& $84.1_{4.9}$ & $86.8_{5.4}$ & $84.8_{6.7}$ \\
PEG~\cite{re_embedding}  
& $84.0_{3.3}$ & $88.1_{2.4}$ & $84.5_{6.8}$ \\
PPEG~\cite{transmil}  
& $84.8_{1.3}$ & $87.1_{2.4}$ & $84.3_{6.8}$ \\ \hline
w/o PROPE   
& $83.8_{3.2}$ & $87.8_{0.5}$ & $83.9_{1.3}$ \\
PROPE   
& $\mathbf{90.4_{0.6}}$ & $\mathbf{95.0_{1.4}}$ & $\mathbf{90.1_{0.8}}$  \\
\bottomrule[\heavyrulewidth]
\end{tabular}
}
\caption
    {
         The comparison results of various positional encoding methods on the TCGA-BRAC dataset.
         The instance feature extractor utilizes the ImageNet-pretrained ResNet50~\cite{resnet}.
         The results reported include the average and standard deviation of five runs.
         The highest performance is highlighted in bold. 
    }
\label{table:a_brac_prope}
\end{table}

As in Tab.~\ref{table:a_brac_prope} and Tab.~\ref{table:a_lung_prope}, the experimental analysis reveals that PROPE achieves superior performance compared to other positional encoding methods across both the TCGA-LUNG and TCGA-BRAC datasets.
Specifically, on the TCGA-LUNG dataset, PROPE achieved an accuracy of $90.4 \pm 0.6\%$, an AUC of $95.0 \pm 1.4\%$, and an F1 score of $90.1 \pm 0.8\%$, outperforming all other methods including Sinusoidal, Alibi, ROPE, 2D ROPE, PEG, and PPEG. 
Similarly, on the TCGA-BRAC dataset, PROPE demonstrated notable improvements with an accuracy of $90.4 \pm 0.6\%$, an AUC of $95.0 \pm 1.4\%$, and an F1 score of $90.1 \pm 0.8\%$.
These results highlight the effectiveness of PROPE in capturing spatial information for whole-slide image analysis tasks. 
The consistent high performance across different metrics and datasets indicates that PROPE's use of normalized polar coordinates effectively addresses the limitations of other positional encoding methods, such as the length extrapolation issue observed in sinusoidal encoding. 
Thus, PROPE not only enhances model performance but also provides a robust solution for handling diverse data types.

\section{Scaling factor $\lambda$ of PROPE}
\label{se:scaling}

We conducted empirical experiments to evaluate the optimal $\lambda$ value for PROPE. The experiments were performed on three distinct datasets: CAMELYON-16, TCGA-LUNG, and TCGA-BRAC. Specifically, we tested five different values of $\lambda$: 1, 512, 2048, 4096, and 8192. Our analysis aimed to identify which value of $\lambda$ would yield the best performance across three key metrics: Accuracy (ACC), Area Under the Curve (AUC), and F1 score.
As illustrated in Table~\ref{table:a_scaling_factor}, setting $\lambda$ to 512 consistently yields the highest performance across all three datasets and metrics. For instance, on the CAMELYON-16 dataset, a $\lambda$ value of 512 achieved an accuracy of $90.3 \pm 0.6\%$, an AUC of $96.1 \pm 1.2\%$, and an F1 score of $89.3 \pm 1.6\%$. Similarly, on the TCGA-LUNG dataset, the same $\lambda$ value resulted in an accuracy of $90.4 \pm 0.6\%$, an AUC of $95.0 \pm 1.4\%$, and an F1 score of $90.1 \pm 0.8\%$. On the TCGA-BRAC dataset, although the performance was relatively lower compared to other datasets, $\lambda = 512$ still provided the best results with an accuracy of $85.4 \pm 4.2\%$, an AUC of $89.1 \pm 0.5\%$, and an F1 score of $84.9 \pm 1.3\%$.
All experiments were conducted using a ResNet50 feature extractor, and the results include the average and standard deviation of five runs. The highest performance for each dataset and metric is highlighted in bold in Table~\ref{table:a_scaling_factor}. These findings indicate that a $\lambda$ value of 512 is optimal for achieving the best overall performance across various datasets and evaluation metrics.
\begin{table*}[t]
\centering
\resizebox{\linewidth}{!}{
\begin{tabular}{c | c  c  c  | c  c  c  | c c c } 
\hline
  \multirow{2}{*}{Methods}         & \multicolumn{3}{c|}{CAMELYON-16}                                          & \multicolumn{3}{c|}{TCGA-LUNG}           &  \multicolumn{3}{c}{TCGA-BRAC}   \\ 
                                   & ACC & AUC &  F1 &  ACC  & AUC & F1 &  ACC  & AUC &  F1 \\ 
  \hline
  $\lambda$ = 1
  &    $89.9_{1.3}$      &  $91.0_{1.3}$          & $89.7_{0.3}$    
  &  $94.8_{1.1}$             & $97.4_{0.5}$                & $95.2_{1.0}$                  
  &   $91.6_{3.0}$	       & $95.4_{2.1}$	 & $89.0_{1.0}$ \\
  $\lambda$ = 512
  & \bm{$90.3_{0.6}$}   & \bm{$96.1_{1.2}$}	       & \bm{$89.3_{1.6}$}	              
  & \bm{$90.4_{0.6}$}	  & \bm{$95.0_{1.4}$}          & \bm{$90.1_{0.8}$}	             
  &  \bm{$85.4_{4.2}$}      & \bm{$89.1_{0.5}$}      & \bm{$84.9_{1.3}$}\\
  $\lambda$ = 2048
  &  $89.8_{1.3}$    &  $95.8_{0.3}$	   & $89.0_{1.1}$                
  &  $90.1_{0.8}$	   &  $94.7_{0.4}$	   & $89.7_{0.8}$  
  & $85.0_{4.0}$	                  &   $88.7_{2.0}$	     &$88.9_{2.6}$	               \\
  $\lambda$ = 4096
  &  $89.7_{3.3}$  &  $95.7_{3.3}$	   & $89.0_{2.2}$                
  &  $90.0_{0.7}$	 &  $94.6_{0.7}$	   & $89.6_{0.6}$  
  & $84.9_{3.3}$	                  &   $88.5_{2.2}$	     &$88.0_{3.7}$	               \\
  $\lambda$ = 8192
 & $89.9_{0.6}$ & $96.0_{1.2}$	       & $89.0_{1.3}$	          
 &  $90.2_{1.4}$	         & $94.7_{2.3}$	            & $89.6_{1.5}$	           
  &   $85.1_{4.6}$	      &$88.8_{1.6}$	               & $88.9_{3.2}$\\
\hline
\end{tabular}
}
\caption
    {
    The performance of different $\lambda$ values evaluated on the CAMELYON-16, TCGA-LUNG, and TCGA-BRAC datasets is reported.
    The instance feature extractor utilizes the ResNet50 model.
    The results include the average and standard deviation of five runs.
    The highest performance is highlighted in bold.
    }
\label{table:a_scaling_factor}
\end{table*}

\section{Visualizaton}
\label{sec:a_vis}

\begin{figure*}[t]
    \centering
    \includegraphics[width=1\linewidth]{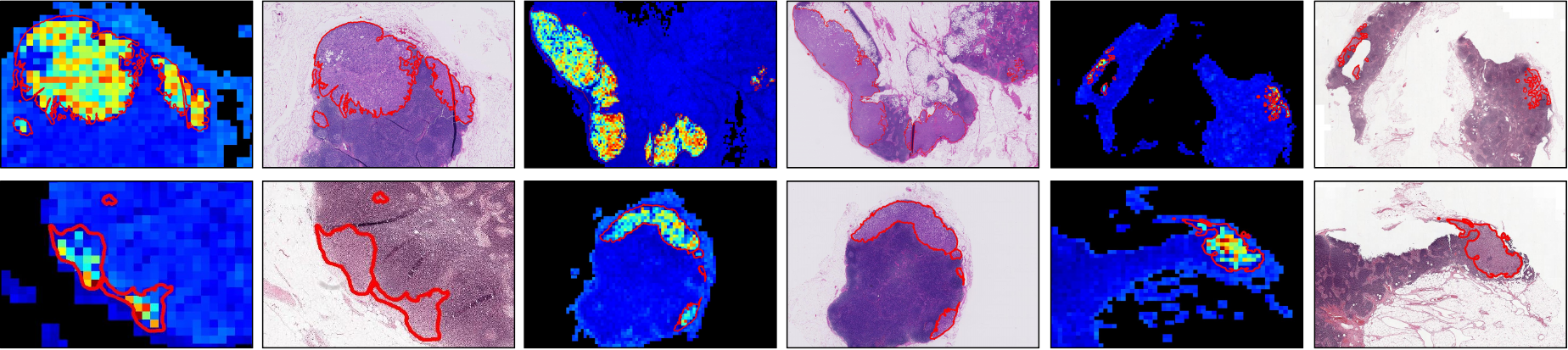}
    \vspace{-0.4em}
    \caption{
        Patch visualization produced by SAC-MIL.
        }
    \label{fig:vis_v2}
    \vspace{-0.9em}
\end{figure*}

In Figure~\ref{fig:vis_v2}, we present the predictions of SAC-MIL on the CAMELYON-16 dataset.
The L2 norm of the outputs from the final MLP layer of SAC-MIL is computed. 
The tumor region is effectively identified, with minimal instances detected outside the tumor boundary. 
This suggests that our method achieves a low false positive rate in detecting tumor instances with high precision.
The distinct separation of tumor boundaries underscores the reliability and precision of our approach.
The majority of tumor regions are accurately pinpointed, showcasing the capability of SAC-MIL in managing intricate tissue architectures.

\end{document}